\documentclass{article}

\usepackage{graphicx}
\usepackage{tabularx}
\usepackage{enumerate}
\usepackage{multirow}
\usepackage{listings}
\usepackage{xcolor}
\usepackage{comment}
\usepackage{amsfonts}
\usepackage{algorithm}
\usepackage{algpseudocode}
\usepackage{pifont}
\usepackage{multirow}
\usepackage{subcaption}
\usepackage{xspace}

\definecolor{codegreen}{rgb}{0,0.6,0}
\definecolor{codegray}{rgb}{0.5,0.5,0.5}
\definecolor{codepurple}{rgb}{0.58,0,0.82}
\definecolor{backcolour}{rgb}{0.95,0.95,0.92}

\lstdefinestyle{mystyle}{
	backgroundcolor=\color{backcolour},   
	commentstyle=\color{codegreen},
	keywordstyle=\color{magenta},
	numberstyle=\tiny\color{codegray},
	stringstyle=\color{codepurple},
	basicstyle=\ttfamily\footnotesize,
	breakatwhitespace=false,         
	breaklines=true,                 
	captionpos=b,                    
	keepspaces=true,                 
	numbers=left,                    
	numbersep=5pt,                  
	showspaces=false,                
	showstringspaces=false,
	showtabs=false,                  
	tabsize=2
}

\makeatletter
\def\blfootnote{\xdef\@thefnmark{}\@footnotetext}
\makeatother

\begin{document}

\title{%
  Sales Channel Optimization via Simulations Based on Observational Data with Delayed Rewards: A Case Study at LinkedIn}

\author{Diana M.  Negoescu \and
Pasha Khosravi$^*$ \and
Shadow Zhao \and
Nanyu Chen$^{**}$ \and
Parvez Ahammad \and
Humberto Gonzalez}

\newcommand{\todo}[1]{\textbf{<{\color{red} ToDo:} #1>}}
\newcommand{\editdn}[1]{\textbf{<{\color{blue} EDIT:} #1>}}
\newcommand{\channeldescriptor}[1]{\emph{#1}\xspace}
\newcommand{\chonline}{\channeldescriptor{A}}
\newcommand{\chfield}{\channeldescriptor{B}}
\newcommand{\chae}{\channeldescriptor{C}}
\newcommand{\sufeature}{\hat{f}}
\newcommand{\R}{\mathbb{R}}

\date{%
	\small{$^*$UCLA\\
	$^{**}$Gopuff\\
	LinkedIn Corporation\\}
	\phantom{}\\
	\today}

\maketitle

\begin{abstract}
Training models on data obtained from randomized experiments is ideal for making good decisions. However, randomized experiments are often time-consuming, expensive, risky, infeasible or unethical to perform, leaving decision makers little choice but to rely on observational data collected under historical policies when training models. This opens questions regarding not only which decision-making policies would perform best in practice, but also regarding the impact of different data collection protocols on the performance of various policies trained on the data, or the robustness of policy performance with respect to changes in problem characteristics such as action- or reward- specific delays in observing outcomes. We aim to answer such questions for the problem of optimizing sales channel allocations at LinkedIn, where sales accounts (leads) need to be allocated to one of three channels, with the goal of maximizing the number of successful conversions over a period of time. A key problem feature constitutes the presence of stochastic delays in observing allocation outcomes, whose distribution is both channel- and outcome- dependent. We built a discrete-time simulation that can handle our problem features and used it to evaluate: a) a historical rule-based policy; b) a supervised machine learning policy (XGBoost); and c) multi-armed bandit (MAB) policies, under different scenarios involving: i) data collection used for training (observational vs randomized); ii) lead conversion scenarios; iii) delay distributions. Our simulation results indicate that LinUCB, a simple MAB policy, consistently outperforms the other policies, achieving a 18-47\% lift relative to a rule-based policy\blfootnote{Contact: dnegoescu@linkedin.com.  Accepted at REVEAL'22 RecSys Workshop}.
\end{abstract}

\section{Introduction}
\label{sec:Randomized}

The problem known as \emph{sales channel optimization} is shared among all organizations reaching out to potential customers, also defined as \emph{leads}, where they must decide the initial communication channel for customer contact (e.g., email, phone, etc.).
Given a fixed set of channels and a pool of leads, the sales channel optimization problem aims to find a suitable balance between cost and an unknown success likelihood of each potential communication, which can be naturally formulated as a multi-armed bandit (MAB)  \cite{sutton2018reinforcement}.
Data-driven policies for this optimization problem can only be built if previously collected historical training data is available, including lead contextual information, channel(s) used to communicate with them,  feedback delay, and whether the lead was converted into a client.
While such a data set is best generated through randomized experiments, in many situations these experiments are risky, expensive or time-consuming, particularly when actions have long delays in providing outcome feedback, or involve extensive labor or financial resources to implement.
In the absence of data from randomized experiments, practitioners often rely on observational data, which can suffer from various forms of bias~\cite{rosenbaum2010design}, resulting in suboptimal performance of models trained on it.
In this paper,  we develop a simulation framework to empirically study the impact of randomized vs.\ observational training data generation, and estimate the performance of MAB and supervised machine learning (ML) policies in a sales channel optimization problem at LinkedIn.

LinkedIn's go-to-market strategy includes designing and implementing policies for the sales channel optimization of our products~\cite{LIProd}.
In particular, in a LinkedIn Sales Solutions product,  we consider three channels:
\begin{itemize}
\item Channel~\chonline: this channel relies on online communications, is highly automated, and thus requires little manual effort. Outcome feedback delay is short, measured in tens of days.
\item Channel~\chfield: this channel relies on personal communication with representatives, with the goal of establishing a long-term relationship. Outcome feedback delay is long, between a few months to a year.
\item Channel~\chae: this channel strikes a balance between the two above, involving less manual effort than Channel~\chfield, but with longer feedback delays than Channel~\chonline.
\end{itemize}

Historically, channel allocations among these three options have been made according to rule-based policies that allocate leads depending on a handful of arbitrarily chosen features, raising concerns that the outcomes data collected through this policy may suffer from selection bias.
Moreover, reward delays in this setting, as well as cost of allocations made via Channels~\chfield and~\chae, result in prohibitive conditions for randomized experiments.

\section{Related Work}
\label{sec:related_work}

Simulation models have been used extensively in many domains, including business, computing, medical, and social sciences~\cite{sokolowski2011principles}.  In particular, the evaluation of optimization policies typically involves either generating synthetic data or attempting to evaluate the policies using previously collected (or logged) data, a problem commonly known as ``off-policy learning''~\cite{levine2020offline}.
When large, uniformly randomized, logged data sets are available, \cite{li2010contextual} shows that unbiased evaluation of a new policy is possible. 
Other policy evaluation techniques for non-uniform randomized logged data sets originated in the causal analysis field~\cite{imbens2015causal, dudik2011doubly, dudik2011efficient, schnabel2016recommendations, joachims2017unbiased, wang2017optimal, huang2020keeping}, which require the hard-to-achieve constraint that policies have coverage of the entire feature space.

Multi-armed bandits are used to tackle exploration-exploitation trade-offs in problems where a choice must be made between multiple actions with unknown rewards~\cite{thompson1933likelihood, berry1985bandit, cesa2006prediction, gittins2011multi, bubeck2012regret, slivkins2019introduction, lattimore2020bandit}.
Contextual bandits~\cite{hazan2007online}, in which the choice is modulated by extra features, have gained popularity especially since the introduction of \emph{LinUCB}~\cite{li2010contextual, krishnamurthy2015contextual, agarwal2014taming}.
Of particular interest to our result are bandits with delayed feedback~\cite{joulani2013online, pike2018bandits, grover2018best, garg2019stochastic, zhou2019learning, arya2020randomized}.
We expand on their work by focusing on empirical evaluation under the practical constraints of sales channel optimization problems at LinkedIn.

\section{Methods}
\label{sec:methods}

\subsection{Problem setup}

We use the contextual multi-armed bandit framework to model our sales channel optimization problem.
Specifically, at time $t$ we are given an account described by a $d$-dimensional \emph{context} feature vector  $x_t \in \mathcal{X} \subset \R^d$.
We need to allocate each account $x_t$ to one of the available \emph{actions} $a_t \in \mathcal{D}$, resulting in a \emph{reward} $r(x_t, a_t) \in \{0, 1\}$, observed after a random delay $\tau(x_t, a_t)$.
We note that the delay time $\tau$ depends on the action and is not independent of the reward $r(x_t, a_t)$.
Our goal is then to maximize expected cumulative rewards observed over a given time horizon $T$: $\mathbb{E}[\sum_{t =1}^T r(x_t, a_t) \mathbf{1}_{\{t+\tau(x_t, a_t) \leq T\}}]$.

In the context of our product of interest at LinkedIn, we typically have to allocate a few hundred leads each day to one of three channels, i.e., $\mathcal{D} = \{\chonline, \chfield, \chae\}$.
Rewards in this case are lead conversions into clients, set to~1 when converted and~0 otherwise, which are executed at time $t$ and observed at time $t + \tau(x_t, a_t)$. 
Thus, our objective is to maximize the total number of lead conversions.

\subsection{Simulation details}
We build a discrete time simulation informed by a data set of leads initiated at LinkedIn between July~1st, 2020 and April~1st, 2021, in order to evaluate the performance of different channel allocation policies under several state-of-the-world scenarios regarding the nature of available training sets (randomized vs.  observational),  reward ground truth and feedback delay distributions.
The original data set consisted of 200,000+ accounts, each encoded as a context vector of 13 post-processed features, together with allocated channels, and binary conversion labels.  All labels in our data set were either 0 or 1.
We used one-hot-encoding for categorical variables, normalized z-scores for real-valued variables capped to the interval $[-5,5]$, and imputed missing real-valued features with the population mean.

\begin{figure}[t]
  \centering
  \includegraphics[width=\linewidth]{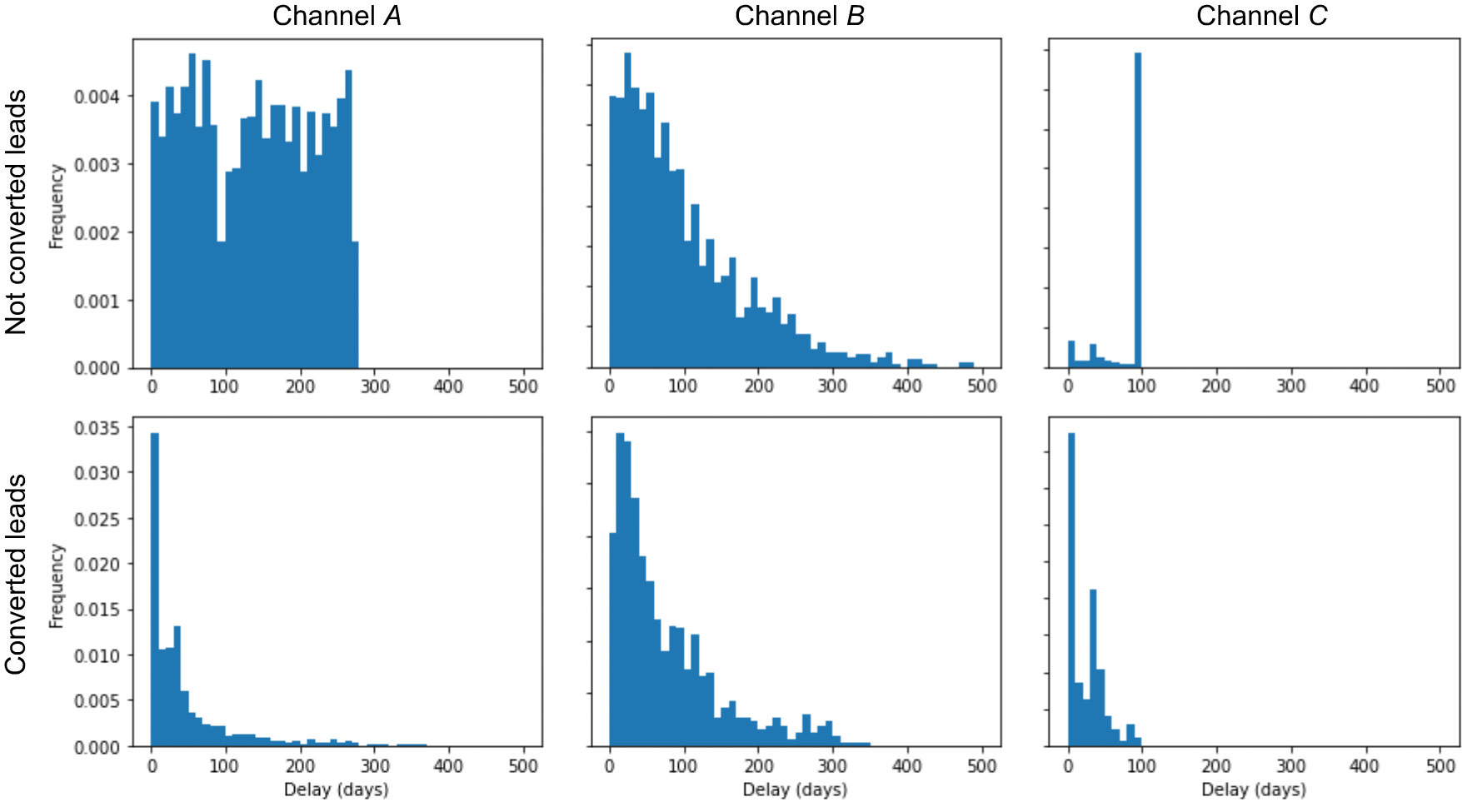}
  \caption{%
    Distribution of outcome delays on the LinkedIn sales channel training data set for each communication channel and conversion status.
    While not converted outcome delays vary widely among different channels, the conversion outcome delays have a more consistent behavior, with Channel~\chonline having the shortest delays on average, and Channel~\chfield the longest.
  }
  \label{fig:delays}
\end{figure}

The simulation tracks daily leads that are allocated according to the user-specified policy (Appendix Algorithm~\ref{simulator}).
 Figure~\ref{fig:delays} shows delay distributions for each channel and conversion status on this data set.

\subsubsection{Logged data set collection scenarios}
\label{subsubsection:pretraining}

First, we simulate different logged data set collection methodologies, which the simulated policy will consume as available data, using the following three scenarios:
\begin{itemize}
\item \textbf{Fully Randomized}: Assign channel allocations uniformly at random.
\item \textbf{Observational}: Assigning channels according to the historically used rule-based policy based on one feature value, denoted $\sufeature$.
  That is, $a_t = \chonline$ if $\sufeature < 10$, $a_t = \chfield$ if $\sufeature > 20$, and $a_t = \chae$ otherwise.

\item \textbf{Partially Randomized}: Randomize by $\sufeature$ values within a range to collect training data,  in order to avoid major deviations from the historically applied rule-based policy. In this case, $a_t = \chonline$ if $\sufeature < 10$, $a_t = \chfield$ if $\sufeature > 20$, and, either $a_t = \chae$ with probability~$\frac{2}{3}$ or $a_t = \chfield$ with probability~$\frac{1}{3}$ otherwise.
\end{itemize}

\subsubsection{Lead conversion scenarios}

To simulate non-trivial, yet representative, lead conversion distribution scenarios, we first used $k$-means clustering (with $k = 10$) to aggregate the accounts in our training data set based on their feature vectors $x_t$ (not including the allocations and outcomes).
We then considered the following lead conversion scenarios:
\begin{itemize}
\item \textbf{Historical}: We assign each cluster-action pair a Bernoulli probability of success, whose parameter $p$ is computed as the MLE on our LinkedIn data set.
\item \textbf{Uniform}: We assign each cluster-action-$\sufeature$ triple a Bernoulli probability of success, whose parameter $p$ is sampled from a uniform distribution over an interval informed by the success probabilities of the Historical scenario.
\item \textbf{$\sufeature$-adjusted uniform}: Similar to the Uniform scenario, but give more importance to $\sufeature$ values by increasing the probability of conversion when the action-$\sufeature$ pair is the same as in the Observational scenario.
\end{itemize}

\subsubsection{Delay scenarios}

We simulated delays by sampling from the population distribution on our LinkedIn data set conditional on action and reward.
On top of the samples, we enabled three scenarios where each resulting simulated delay is multiplied by a constant factor $\lambda_{delay} \in \{0, \frac{1}{2}, 1\}$.

\subsubsection{Policy simulation}

Finally, we implemented the following LinkedIn-relevant policies using our simulator:
\begin{itemize}
\item \textbf{Rule-based $\sufeature$ policy}: Rule-based channel assignment policy identical to the Observational scenario in~\ref{subsubsection:pretraining}.
\item \textbf{XGBoost supervised ML}: We train an \emph{XGBoost}~\cite{chen2016xgboost} model using simulated logged data from~\ref{subsubsection:pretraining}, and choose the action with the highest predicted probability of conversion.
  The simulator retrains the model every 90 days.
\item \textbf{$\epsilon$-greedy MAB}: As described in~\cite{sutton2018reinforcement}, $\epsilon$ percent of the time we allocate a channel uniformly at random, the rest of the time follow the same recommendation as the XGBoost supervised ML policy,  retraining every 90 days.
\item \textbf{LinUCB MAB}: As described in~\cite{li2010contextual}, we choose the channel with highest upper-confidence bound (predicted mean + confidence interval) as defined by a linear relationship between features and conversion.
  Updates the upper-confidence bound online as new data becomes available.
\end{itemize}

\section{Results}
\label{sec:results}

\begin{figure}[t]
  \centering
  \includegraphics[width=\linewidth]{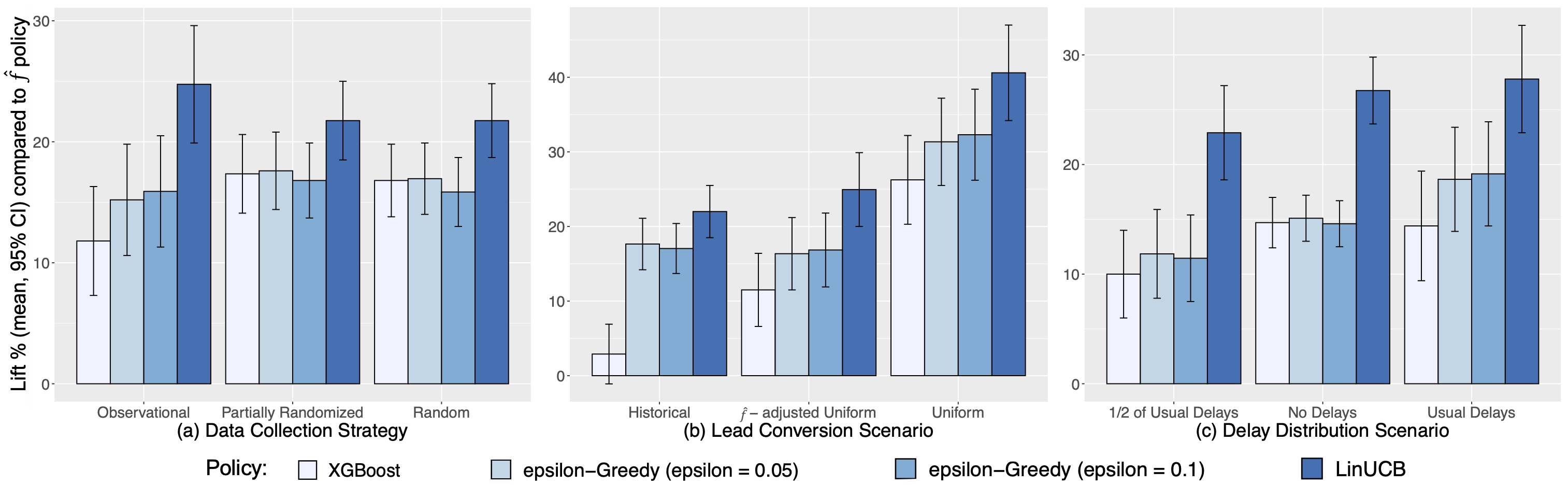}
  \caption{Lift \% with respect to $\sufeature$ policy (100 runs, 95\% CI). Data collection was simulated for 90 days, simulation horizon was one year. Unless varied, data collection was simulated using the \textit{Observational} scenario, lead conversion was according to the $\sufeature$-adjusted Uniform scenario, and delays were simulated using historical data.}
  \label{fig:results}
\end{figure}

Results from our simulation are shown in Figure~\ref{fig:results} (numerical values in Appendix Table~\ref{table:tabular_data}).
We measure cumulative reward lift (as percentage improvement) relative to the historically used rule-based $\sufeature$ policy as estimated by our simulator (Appendix Algorithm \ref{simulator}).
We found that LinUCB outperforms XGBoost and $\epsilon$-greedy under most simulated scenarios ($18\%$ to $47\%$ lift compared to the ruled-based $\sufeature$ policy), and especially more so when the data used for training was observational (Figure~\ref{fig:results}a).
These results were robust to changes in reward distribution (Figure~\ref{fig:results}b) and delay distribution shifts (Figure \ref{fig:results}c), regardless of the choice of policy.
Moreover, even partial randomization improves the performance of supervised ML policies, but MABs performed well whether or not the training data was collected under a randomized policy.

\section{Acknowledgments}
  We particularly thank
  Justin Dyer,
  Adrian Rivera Cardoso,
  Jilei Yang,
  Ryan Rogers,
  Saad Eddin Al Orjany,
  Wenrong Zeng,
  Yu Liu,
  Rahul Todkar,
  Neha Gupta,
  Maggie Zhang and
  Jennifer Kloke
  for their helpful comments and feedback.

\bibliographystyle{acm}
\bibliography{LSS_simulator_reference}

\newpage

\appendix

\section{Appendix: Simulator algorithm for sales channel allocation}
\begin{algorithm}
  \caption{Channel allocation simulator}
  \label{simulator}
  \begin{algorithmic}[1]
    \Require
\\
    Number of days to simulate $T$;\\
    Channel allocation policy $\pi$  (except for  $\hat{f}$,  this policy is trained on data collected using scenario $\pi_0$);\\
    Lead conversion scenario $L$;\\
    Feature distribution from training data $F(X)$;\\
    Delay distribution from training data per action and reward $\Delta = \{\Delta(a, r)\}_{a \in \mathcal{D}, r \in \{0,1\}}$.
    \Procedure{Simulate($\pi$, $L$, $F(X)$, $\Delta$)}{}
    \State Initialize reward scheduler \texttt{pq}.
    \For{each day in 1 to $T$}:
    \State \texttt{today's leads} $\leftarrow$ Sample from $F(X)$.
    \For{$x_t \in$ \texttt{today's leads}}:
        \For{$a_t \in \mathcal{D}$}:
        \State Sample reward $r(x_t, a_t) \sim L$.
        \State Sample delay $\tau(x_t, a_t) \sim \Delta(a_t, r(x_t, a_t))$.
        \EndFor
    \State Take action $a^{\pi}_t$ according to the policy to be evaluated $\pi$.
    \State Schedule reward $r(x_t, a^{\pi}_t)$ to be observed after $\tau(x_t, a^{\pi}_t)$ days.
    \EndFor
    \State Observe today's scheduled rewards and accumulate total rewards.
    \State Update policy $\pi$ using today's observations (if necessary).
    \EndFor
    \EndProcedure
  \end{algorithmic}
\end{algorithm}

A high level description of our policy simulator can be found in Algorithm \ref{simulator}.  We note that the same algorithm is run prior to the evaluation stage with $\pi = \pi_0$ in order to generate the logged data according to the data set collection scenarios in Section \ref{subsubsection:pretraining},  where policy $\pi_0$ in that case is one of the three scenarios: \textit{Fully Randomized},  \textit{Observational} or \textit{Partially Randomized}.  
\newpage
\section{Appendix: Results in tabular form}
Table~\ref{table:tabular_data} shows the results, plotted in Figure~\ref{fig:results}, in tabular form.
\begin{table}[ht]
  \begin{subtable}{\linewidth}
    \centering
    \begin{tabular}{|c|c||c|c|c|c|c|}
      \hline
      \multicolumn{2}{|c|}{} & \multicolumn{3}{|c|}{\textbf{Data collection scenario}} \\
      \cline{3-5}
      \multicolumn{2}{|c|}{} & \textit{Observational} & \textit{Partially randomized} & \textit{Fully Random} \\
      \hline
      \multirow{5}{*}{\rotatebox{90}{\textbf{Policy}}}  & Rule-based $\sufeature$ & [0.0, 0.0] & [0.0, 0.0] & [0.0, 0.0] \\
      \cline{2-5}
                             & \emph{LinUCB} & [19.9, 29.6] & [18.5, 25.0]  & [18.7, 24.8] \\
      \cline{2-5}
                             & \emph{XGBoost} & [7.3, 16.3] & [14.1, 20.6]  & [13.8, 19.8] \\
      \cline{2-5}
                             & \emph{$\epsilon$-greedy} ($\epsilon = 0.05$) & [10.6, 19.8] & [14.4, 20.8] & [14.0, 19.9] \\
      \cline{2-5}
                             & \emph{$\epsilon$-greedy} ($\epsilon = 0.1$) & [11.3, 20.5] & [13.7, 19.9] & [13.0, 18.7] \\
      \hline
    \end{tabular}
    \caption{Varying data collection: Lift \% wrt rule-based $\sufeature$ policy (100 runs, 95\% CI).}
    \label{table:data_collection}
  \end{subtable}
  \begin{subtable}{\linewidth}
    \centering
    \begin{tabular}{|c|c||c|c|c|c|c|}
      \hline
      \multicolumn{2}{|c|}{} & \multicolumn{3}{|c|}{\textbf{Lead conversion scenario}} \\
      \cline{3-5}
      \multicolumn{2}{|c|}{} & \textit{Historical} & \textit{Uniform} & \textit{$\sufeature$-adjusted Uniform} \\
      \hline
      \multirow{5}{*}{\rotatebox{90}{\textbf{Policy}}}  & Rule-based $\sufeature$ & [0.0, 0.0] &  [0.0, 0.0] &  [0.0, 0.0]  \\
      \cline{2-5}
                             & \emph{LinUCB} & [18.5, 25.5] & [34.2, 47.0]  & [20.0, 29.9] \\
      \cline{2-5}
                             & \emph{XGBoost} & [-1.1, 6.9] & [20.3, 32.2]  & [6.6, 16.4] \\
      \cline{2-5}
                             & \emph{$\epsilon$-greedy} ($\epsilon = 0.05$) & [14.2, 21.1] & [25.5, 37.2] & [11.5, 21.2] \\
      \cline{2-5}
                             & \emph{$\epsilon$-greedy} ($\epsilon = 0.1$) & [13.7, 20.4] & [26.2, 38.4] & [11.9, 21.8] \\
      \hline
    \end{tabular}
    \caption{Varying lead conversion: Lift \% wrt rule-based $\sufeature$ policy (100 runs, 95\% CI).}\label{table:lead_distribution}
  \end{subtable}
  \begin{subtable}{\linewidth}
    \centering
    \begin{tabular}{|c|c||c|c|c|c|c|}
      \hline
      \multicolumn{2}{|c|}{} & \multicolumn{3}{|c|}{\textbf{Delay distribution scenario}} \\
      \cline{3-5}
      \multicolumn{2}{|c|}{} & \textit{No Delays} & \textit{1/2 of Usual Delays} & \textit{Usual Delays} \\
      \hline
      \multirow{5}{*}{\rotatebox{90}{\textbf{Policy}}} & Rule-based $\sufeature$ & [0.0, 0.0] &  [0.0, 0.0]&  [0.0, 0.0]  \\
      \cline{2-5}
                             & \emph{LinUCB} & [23.7, 29.8] & [18.6, 27.2]  & [22.9, 32.7] \\
      \cline{2-5}
                             & \emph{XGBoost} & [12.4, 17.0] & [6.0, 14.0]  & [9.4, 19.4] \\
      \cline{2-5}
                             & \emph{$\epsilon$-greedy} ($\epsilon = 0.05$) & [13.0, 17.2] & [7.8, 15.9] & [13.9, 23.4] \\
      \cline{2-5}
                             & \emph{$\epsilon$-greedy} ($\epsilon = 0.1$) & [12.5, 16.7] & [7.5, 15.4] & [14.4, 23.9]\\
      \hline
    \end{tabular}
    \caption{Varying delay: Lift \% wrt rule-based $\sufeature$ policy (100 runs, 95\% CI).}\label{table:delays}
  \end{subtable}
  \caption{Results after changing simulation scenarios as described in Section~\ref{sec:results}.}
  \label{table:tabular_data}
\end{table}

\end{document}